%% file: samplepaper.tex
\documentclass[runningheads]{llncs}
\usepackage[T1]{fontenc}
\usepackage[utf8]{inputenc}
\usepackage{cite}
\usepackage{graphicx}
\usepackage{xcolor}
\usepackage{booktabs}
\usepackage{multirow}
\usepackage{amsmath,amssymb}
\usepackage{url}
\usepackage[colorlinks=true, linkcolor=blue, citecolor=blue, urlcolor=blue]{hyperref}
\usepackage{subcaption}

\newcommand{\dataset}{\mbox{JaWildText}}
\renewcommand{\paragraph}[1]{\noindent\textbf{#1}\hspace{6pt}}

\input{tables/dataset_vars.tex}

\begin{document}

\title{JaWildText: A Benchmark for Vision-Language Models on Japanese Scene Text Understanding}

\titlerunning{JaWildText: Japanese Scene Text Understanding Benchmark}

\author{
Koki Maeda\inst{1,2}\orcidID{0009-0008-0529-3152} \and
Naoaki Okazaki\inst{1,2}\orcidID{0000-0001-7635-6175}
}

\authorrunning{K. Maeda and N. Okazaki}

\institute{Institute of Science Tokyo, Tokyo, Japan \and
Research and Development Center for Large Language Models, National Institute of Informatics, Tokyo, Japan \\
\email{\{koki.maeda@nlp.,okazaki@\}comp.isct.ac.jp}}

\maketitle

\begin{abstract}

Japanese scene text poses challenges that multilingual benchmarks often fail to capture, including mixed scripts, frequent vertical writing, and a character inventory far larger than the Latin alphabet.
Although Japanese is included in several multilingual benchmarks, these resources do not adequately capture the language-specific complexities.
Meanwhile, existing Japanese visual text datasets have primarily focused on scanned documents, leaving in-the-wild scene text underexplored.
To fill this gap, we introduce \textbf{\dataset}, a diagnostic benchmark for evaluating vision-language models (VLMs) on Japanese scene text understanding.
\dataset{} contains \JWTotalSamples{} instances from \JWTotalImages{} images newly captured in Japan, with 1.12 million annotated characters spanning \JWUniqueChars{} unique character types. 
It comprises three complementary tasks that vary in visual organization, output format, and writing style: (i) Dense Scene Text Visual Question Answering (STVQA), which requires reasoning over multiple pieces of visual text evidence; (ii) Receipt Key Information Extraction (KIE), which tests layout-aware structured extraction from mobile-captured receipts; and (iii) Handwriting OCR, which evaluates page-level transcription across various media and writing directions.
We evaluate 14 open-weight VLMs and find that the best model achieves an average score of 0.64 across the three tasks. 
Error analyses show recognition remains the dominant bottleneck, especially for kanji.
\dataset{} enables fine-grained, script-aware diagnosis of Japanese scene text capabilities, and will be released with evaluation code.

\end{abstract}

\begin{figure}[t]
  \centering
  \includegraphics[width=\linewidth]{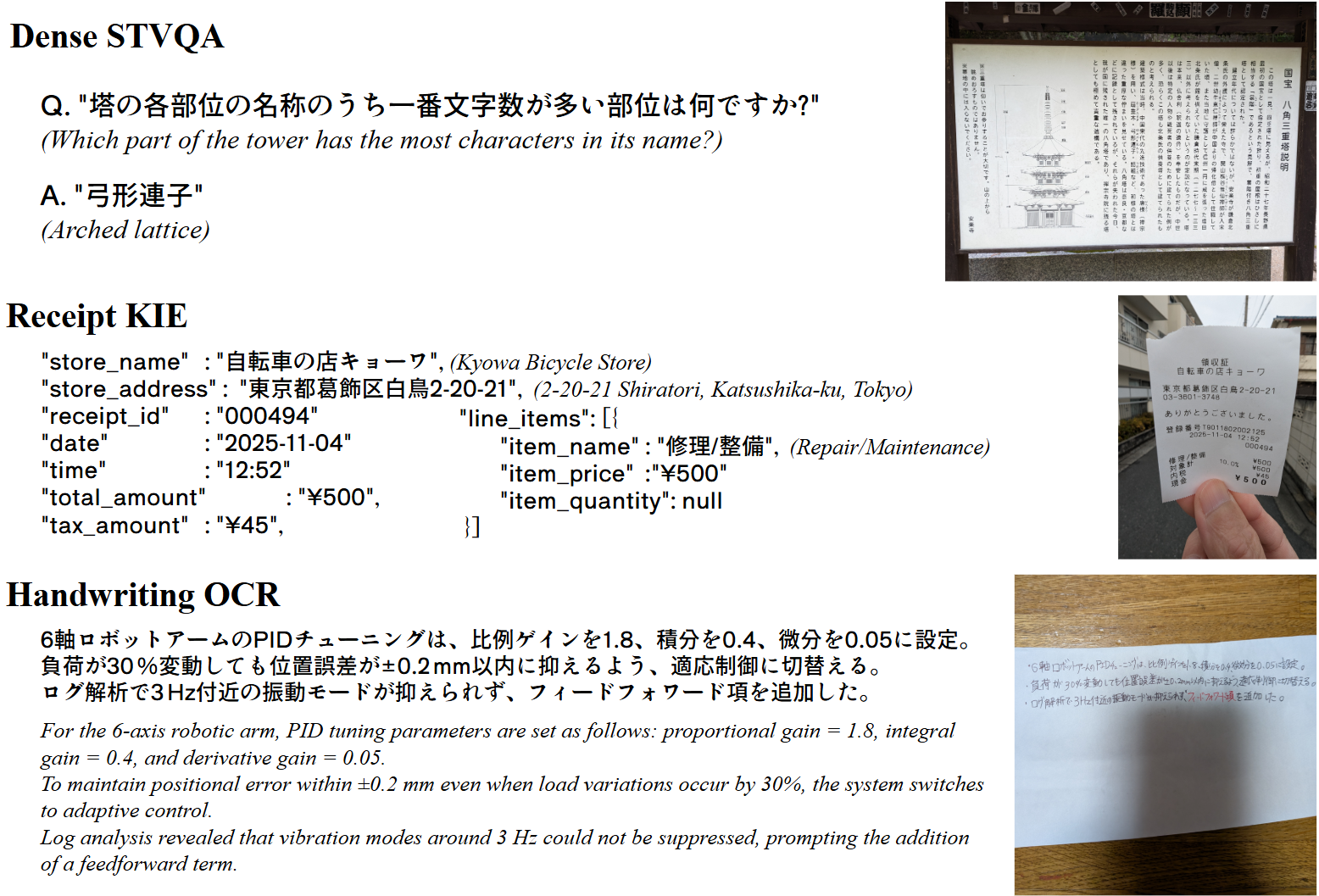}
  \caption{Overview of the \dataset{} benchmark: (i) Dense STVQA, (ii) Receipt KIE, and (iii) Handwriting OCR. We added English translations for readability.}
  \label{fig:overview}
\end{figure}

\section{Introduction}

Text is ubiquitous in everyday environments: on street posters, handwritten notes, receipts, and storefronts.
For decades, text-centric vision systems relied on a modular workflow that first applied OCR to convert pixels into characters and then fed the text to separate modules for downstream tasks~\cite{long-etal-2021-scene}. With the rise of vision-language models (VLMs) such as GPT-4V~\cite{openai-2023-gpt4}, this workflow is shifting toward an end-to-end approach: VLMs generate outputs of a downstream task directly from natural images. This new workflow is increasingly adopted as a practical alternative to dedicated OCR pipelines.

This shift, however, complicates evaluation.
When a VLM is evaluated only by downstream task accuracy, it is often unclear whether an error stems from a failure of character recognition or from incorrect reasoning over correctly recognized text.
Disentangling these two failure modes is critical because precise reading is a prerequisite for any higher-level text understanding. 
Benchmarks that assess reading in natural images and separate recognition failures from reasoning failures are therefore essential, yet few existing resources offer this diagnostic capability, particularly for non-English scripts.

Although multilingual benchmarks~\cite{nayef-etal-2017-rrc-mlt-2017,nayef-etal-2019-rrc-mlt-2019,tang-etal-2024-mtvqa} have extended scene text evaluation beyond English~\cite{singh-etal-2019-textvqa,biten-etal-2019-st-vqa,mathew-etal-2021-docvqa}, they prioritize language breadth over language-specific diagnostics.
This matters particularly for Japanese, whose scene text frequently mixes kanji, hiragana, katakana, and Latin alphanumerics. 
Because the resulting failure patterns differ from those of other scripts, it is difficult to diagnose failures without targeted evaluation.
Existing Japanese scene text resources provide recognition data at the character or word level~\cite{aist-2014-etl-character-db,goto-2020-jpsc1400,rrc-omni-video-2019}. In contrast, Japanese VLM benchmarks target scanned documents or knowledge-centric multimodal tasks~\cite{onami-etal-2024-jdocqa,onohara-etal-2025-jmmmu} without explicitly measuring the underlying recognition ability.
As a result, no existing benchmark can tell whether VLM errors on Japanese scene text arise from recognition or from reasoning.

To fill this gap, we introduce \dataset{}, a fine-grained benchmark for evaluating VLMs on Japanese scene text understanding. The benchmark is designed to disentangle reading from reasoning.
As shown in Figure~\ref{fig:overview}, \dataset{} consists of three tasks that form a compact yet comprehensive configuration. 
The tasks are chosen to vary three factors that commonly confound end-to-end evaluation: visual organization (from cluttered to structured), output format (from free-form to verbatim), and writing style (printed or handwritten).
This design exposes failure modes that remain conflated when models are scored only by downstream task accuracy.
Dense Scene Text Visual Question Answering (Dense STVQA) tests multi-region reading and cross-reference reasoning in cluttered signboards and posters.
Receipt Key Information Extraction (Receipt KIE) evaluates layout-aware structured extraction from in-the-wild imagery.
Handwriting OCR (page-level transcription) assesses long-context transcription of handwritten text, providing a recognition-dominant setting that complements the reasoning-heavy tasks above.
\dataset{} contains \JWTotalSamples{} evaluation instances from \JWTotalImages{} newly collected images in Japan, with 1.12~million annotated characters spanning \JWUniqueChars{} unique characters.

We benchmark 14 open-weight VLMs on \dataset{}. 
The experiments show that the best model achieves an average score of 0.64 across the three tasks.
Our error analysis identified distinct bottlenecks: models that read text accurately may still fail at reasoning, and recognition difficulty varies drastically by script type.
In summary, these results demonstrate that \dataset{} provides fine-grained diagnostic evidence that is invisible in aggregated accuracy alone.

The contributions of this paper are as follows:
\begin{enumerate}
    \item We introduce \dataset{}, to our knowledge, the first benchmark dedicated to evaluating VLMs on Japanese scene text understanding across three complementary tasks grounded in real-world images.
    \item We benchmark 14 open-weight VLMs, establish reproducible baselines, and quantify substantial performance gaps across architectures.
    \item We provide an error analysis that disentangles recognition from reasoning failures, showing that their relative severity varies markedly across model families.
\end{enumerate}

\section{Related Work}

\subsection{Benchmarking Text Understanding in Natural Images}

For English, evaluation resources have matured along three complementary tracks.
\emph{Scene text} benchmarks first targeted detection and recognition in natural images~\cite{yao-etal-2015-icdar15-incidental,veit-etal-2016-cocotext}, and then advanced to text-centric VQA~\cite{singh-etal-2019-textvqa,biten-etal-2019-st-vqa,mathew-etal-2021-docvqa,mathew-etal-2022-infographicvqa,vanlandeghem-etal-2023-dude}, which requires models to read and reason over recognized text.
\emph{Receipt and document understanding} benchmarks such as SROIE~\cite{huang-etal-2021-sroie}, CORD~\cite{park-etal-2019-cord}, and FUNSD~\cite{jaume-etal-2019-funsd} evaluate structured key information extraction~(KIE), testing whether models can map visually organized fields to predefined categories.
\emph{Handwriting recognition} benchmarks, anchored by the IAM Handwriting Database~\cite{marti-etal-2002-iam}, assess verbatim transcription of diverse writing styles; recent work shows that VLM performance degrades substantially on non-English handwriting~\cite{crosilla-etal-2025-benchmark-htr}.
Together, these tracks span a range of visual organization, output format, and writing style, forming a comprehensive evaluation ecosystem for English.

Several benchmarks extend this ecosystem to other languages, progressively broadening language coverage and task complexity.
The ICDAR MLT challenges~\cite{nayef-etal-2017-rrc-mlt-2017,nayef-etal-2019-rrc-mlt-2019} introduce multilingual scene text detection and recognition across up to ten languages. XFUND~\cite{xu-etal-2022-xfund} extends form understanding to seven languages. 
Targeting VLMs directly, MTVQA~\cite{tang-etal-2024-mtvqa} shifted the focus to multilingual text-centric VQA with native annotations. OCRBench~\cite{liu-etal-2024-ocrbench,fu-etal-2025-ocrbenchv2} and CC-OCR~\cite{yang-etal-2024-ccocr} broaden the scope to multilingual OCR for VLMs. 
While these efforts increase language coverage, they treat each language as one among many and provide limited diagnostic depth for language-specific challenges.

Among CJK languages, dedicated benchmarks have emerged for Chinese scene text recognition~\cite{chen-etal-2022-benchmarking-ctr} and Korean text-centric VQA~\cite{hwang-etal-2025-kreta}.
However, Japanese remains without a comprehensive evaluation despite its unique challenges, notably concurrent use of multiple scripts within a single text, complex layouts, and thousands of distinct characters.

\subsection{Japanese Text Understanding}

Existing Japanese-specific resources address isolated facets of text understanding.
For scene text recognition, existing resources target scene text spotting in omnidirectional video~\cite{iwamura-etal-2016-dost}, isolated character classification~\cite{goto-2020-jpsc1400}, vertical text recognition~\cite{sasagawa-etal-2025-vertical-japanese}, and comics~\cite{baek-etal-2026-manga}, restricted to a specific visual setting or textual granularity.
For receipt understanding, existing resources support training or fine-tuning on mobile-captured receipts and post-OCR correction~\cite{nathan-etal-2025-japanese-mobile-receipt-ocr,fujitake-2024-japoc}, but none serve as a benchmark for assessing general-purpose VLM capabilities.
For handwriting, existing datasets provide isolated characters or online stroke data~\cite{aist-2014-etl-character-db,matsumoto-etal-2001-nakayosi,nakagawa-etal-2004-online-jpdb,matsushita-etal-2014-kondate}, or target classical cursive~\cite{clanuwat-etal-2018-kuzushiji}; none covers page-level offline recognition of modern handwriting.
On the reasoning side, JDocQA~\cite{onami-etal-2024-jdocqa} addresses question answering over scanned documents, and JMMMU~\cite{onohara-etal-2025-jmmmu} benchmarks multimodal understanding centered on cultural and academic knowledge rather than text recognition ability.

Mapping these resources onto the three evaluation dimensions of visual organization, output format, and writing style reveals that none connect recognition with reasoning for Japanese scene text.
\dataset{} fills this gap with three complementary tasks that systematically vary these dimensions, enabling fine-grained diagnosis of where and why current VLMs fail on Japanese text in real-world images.

\section{Dataset: \dataset}
\label{sec:dataset}

\dataset{} is designed to expose where a model fails, whether in character recognition, layout understanding, or reasoning, rather than reporting only aggregated task accuracy.
To this end, it comprises three complementary tasks, each annotated to disentangle recognition errors from reasoning and formatting errors.
Because such fine-grained diagnosis requires image diversity and annotation quality that are unavailable in web-scraped corpora, we collected original images and annotations tailored for this work.

\subsection{Dense STVQA}
\label{sec:dense_stvqa}

Dense STVQA evaluates whether a model can read and reason over dense Japanese scene text, using visually complex real-world images such as signboards, bulletin boards, posters, and product packages.

\paragraph{Image Collection.}
To test recognition under realistic conditions, we asked workers from a data collection agency in Japan to photograph text-rich scenes with cameras and smartphones, resulting in \JWBvqaImages{} images. 
We instructed workers to cover indoor and outdoor locations under both daytime and nighttime lighting and avoid multiple shots of the same subject, ensuring diversity in layout, font style, and visual context.
We retained natural artifacts, such as background clutter, partial occlusion, and reflections, to test recognition robustness.

\paragraph{Annotation.}
Annotations are structured into two layers to separate recognition from reasoning.
In the first layer, annotators marked text regions with quadrilateral bounding boxes. 
They transcribed each region, which is defined as a line-level or column-level sequence of visually recognizable characters.
Each image contains \JWBvqaRegionsPerImage{} annotated text regions on average.
In the second layer, native Japanese speakers authored open-ended question-answer pairs over these transcribed regions.
Annotators were encouraged to write questions that require reasoning across multiple text regions rather than extracting a single string from a single area.
We exclude yes/no and multiple-choice formats to minimize chance-level correctness.
Crucially, each question is linked to \emph{evidence regions}: the minimal set of text regions necessary and sufficient to derive the answer.
This linkage enables automatic diagnosis; if a model fails a question but correctly recognizes the evidence regions, the error is attributable to reasoning rather than recognition.
In total, we created \JWBvqaSamples{} question-answer pairs.

\subsection{Receipt KIE}
\label{sec:receipt_kie}

Receipt KIE evaluates structured field extraction from real-world photographs of Japanese receipts. 
This setting introduces challenges largely absent from scene text: rigid columnar layouts, mixed use of full-width and half-width characters, and domain-specific abbreviations produced by thermal printers.

\paragraph{Image Collection.}
Diagnostic value depends on testing under realistic capture conditions; hence, we collected photographs of consumer receipts from everyday transactions rather than flatbed scans.
We retained natural artifacts such as creases, folds, and hand-held tilt to ensure that models are evaluated against the geometric and photometric distortions encountered in practical use.
To maximize visual diversity, we disallowed multiple receipts from the same store while permitting receipts from different branches of the same chain, yielding \JWRkImages{} unique receipt images.

\paragraph{Annotation.}
We annotate receipts at the individual field level so that evaluation can pinpoint which field types a model struggles with, rather than producing only a single per-receipt score.
Our key schema builds on the four header fields of SROIE~\cite{huang-etal-2021-sroie} (\texttt{store\_name}, \texttt{date}, \texttt{store\_address}, \texttt{total\_amount}) and extends it in two directions tailored to Japanese receipts: three additional header fields (\texttt{receipt\_id}, \texttt{time}, \texttt{tax\_amount}) that are usually printed but absent from SROIE, and line-item-level tuples of \texttt{item\_name}, \texttt{item\_price}, and \texttt{item\_quantity}, which test a model's ability to maintain structured alignment across repeated rows.
For each field, annotators recorded the text string and a quadrilateral bounding box; fields absent from a receipt are explicitly marked as null, allowing evaluation to distinguish extraction errors from correct recognition of absence.
In total, we annotated \JWRkRegions{} text regions across \JWRkSamples{} receipts.

\subsection{Handwriting OCR}
\label{sec:handwriting_ocr}

Handwriting OCR evaluates page-level recognition of multi-sentence handwritten Japanese text. Unlike isolated-word or single-line recognition, page-level evaluation requires models to handle layout interpretation, line segmentation, and script mixing simultaneously, reflecting realistic reading scenarios.

\paragraph{Image Collection.}
To obtain diverse yet controlled handwriting samples, we designed a collection pipeline that separates content generation from handwriting production.
First, we defined over 100 genre-keyword pairs spanning everyday topics such as work planning, travel, and cooking. We generated up to 20 prompt texts per pair using a large language model.\footnote{Specifically, we used \texttt{openai/gpt-oss-120b} to generate the prompt texts. The generated texts serve only as writing prompts.}
Each prompt was constrained to approximately 100 characters, long enough to span multiple lines across mixed scripts yet short enough to fit naturally on a single page.

Then, we distributed the prompts to \JWHwWriters{} native Japanese writers, who transcribed them onto designated media and photographed the results using their own devices.
To systematically introduce visual variation, we specified the writing medium and writing direction (horizontal or vertical) for each instance. 
The media include lined paper, unlined plain paper, whiteboards, and tablets, while writers choose line-break positions freely.
Each instance is accompanied by metadata recording the writer ID, writing medium, writing instrument, ink color, and writing direction, enabling fine-grained analysis of how these factors affect recognition performance.

\paragraph{Annotation.}
For each image, the target output is a transcription of all visible handwritten text.
Annotators transcribed the text line by line and drew a quadrilateral bounding box around each region.
A key design principle is that the ground truth should reflect what is visually present in the image rather than what the writer intended.
Accordingly, we preserved writer-introduced errors such as misspellings or omitted characters. 
For characters that a writer started but left incomplete due to writing errors, we assigned a dedicated symbol ($\square$) to mark them explicitly.
This annotation policy ensures that model evaluation measures visual recognition fidelity rather than error correction ability.
We annotated \JWHwSamples{} handwriting instances with \JWHwRegions{} lines, totaling \JWHwChars{} characters.

\input{tables/stats_overall}

\begin{figure*}[t]
  \centering
  \includegraphics[width=\textwidth]{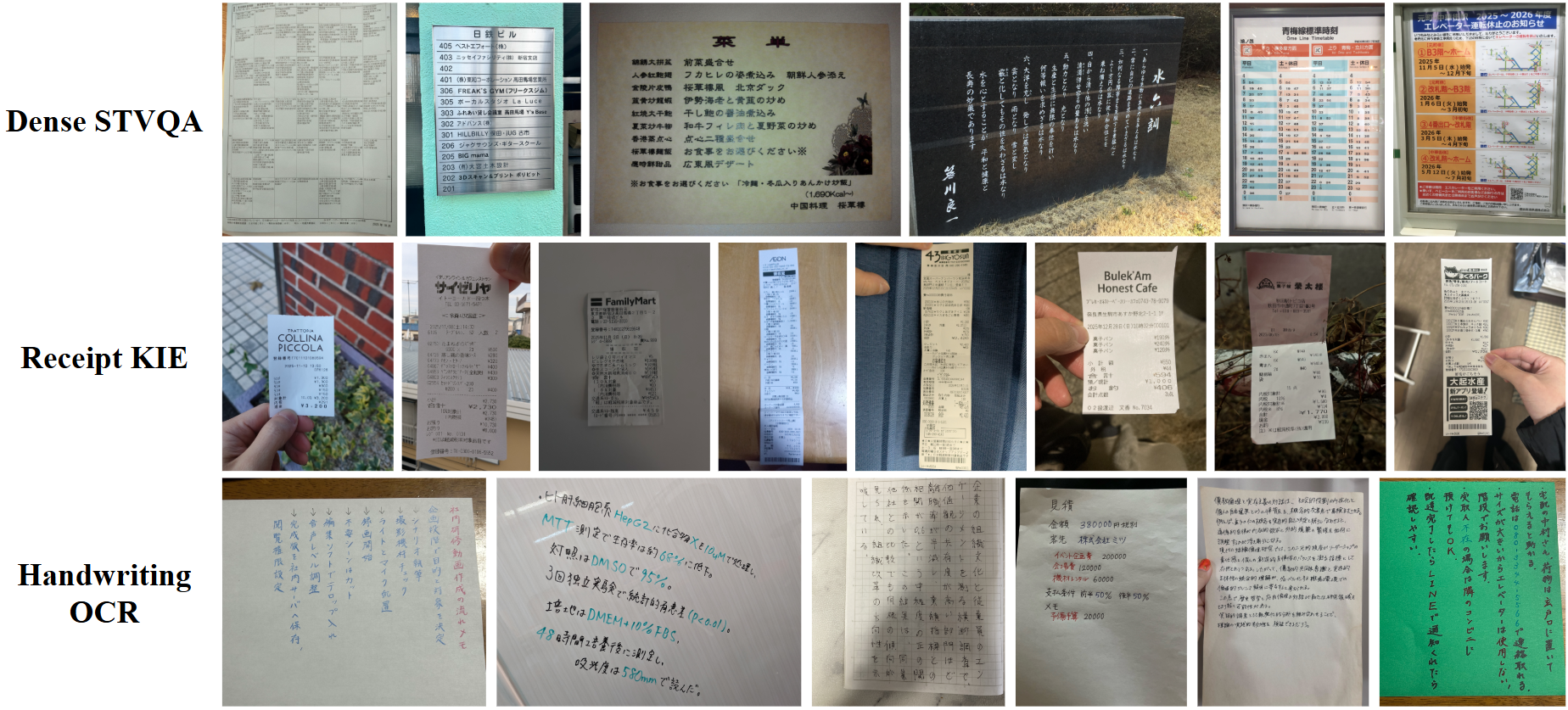}
  \caption{Representative images from each task, illustrating the diversity of \dataset{}. Dense STVQA covers signboards, posters, and product packages under varying conditions. Receipt KIE includes receipts with creases, folds, and diverse perspectives. Handwriting OCR spans multiple writing media and directions.}
  \label{fig:sample_images}
\end{figure*}

\begin{figure}[t]
    \centering
    % ====================================
    \begin{subfigure}[t]{0.35\linewidth}
        \centering
        \includegraphics[width=\linewidth]{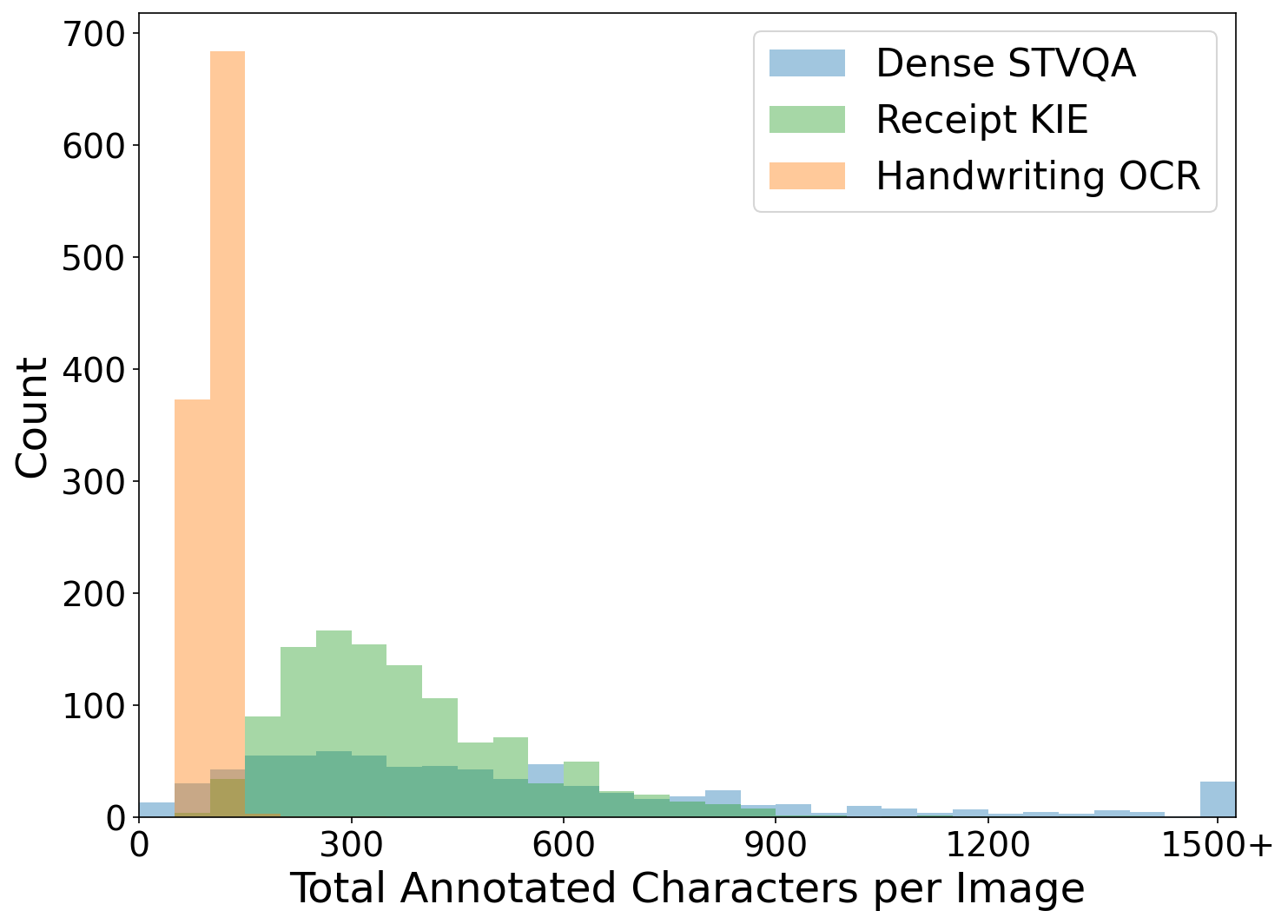}
        \caption{Distribution of total character length per image.}
        \label{fig:char_density}
    \end{subfigure}
    % ====================================
\hfill
    % ====================================
    \begin{subfigure}[t]{0.63\linewidth}
        \centering
        \includegraphics[width=\linewidth]{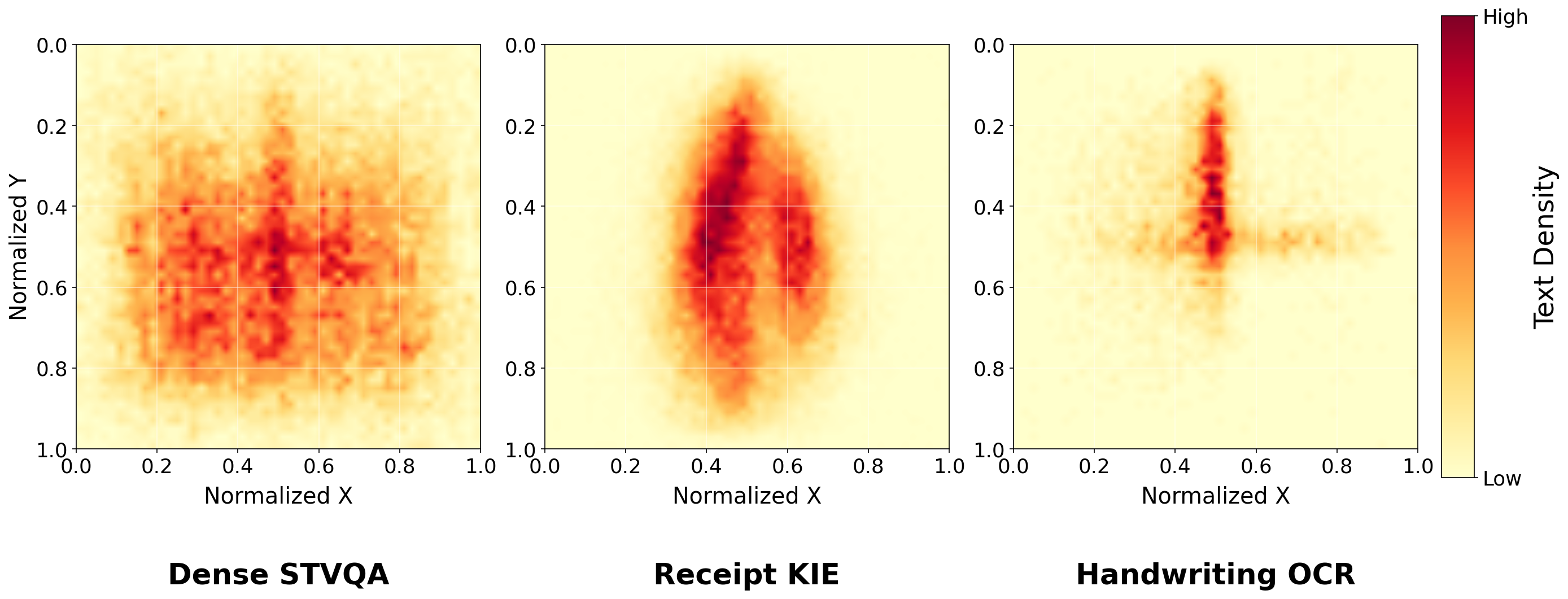}
        \caption{Spatial distribution of text-region centers.}
        \label{fig:text_heatmap}
    \end{subfigure}
    % ====================================
    \caption{Image-level text properties in \dataset.}
    \label{fig:image_text_properties}
\end{figure}

\subsection{Quality Control}
\label{sec:quality_control}

Image collection and annotation were conducted by a professional data curation agency with compensated annotators.
To calibrate annotation guidelines before full-scale production, the authors and the agency jointly reviewed the first 10\% of deliverables and refined the guidelines based on observed inconsistencies.
In the main phase, each instance was labeled by one annotator and independently verified by a second; disagreements were resolved through discussion.
We excluded images containing non-public personally identifiable information, such as faces, vehicle license plates, or credit card numbers, while retaining publicly displayed information (e.g., store phone numbers on receipts) needed for the benchmark tasks.
The dataset, including all images and annotations, will be publicly released under the Apache License 2.0.\footnote{\url{https://huggingface.co/datasets/llm-jp/jawildtext}}

\subsection{Dataset Statistics}
\label{sec:statistics}
Table~\ref{tab:dataset_stats} summarizes key statistics of \dataset{}.
The dataset comprises \JWTotalSamples{} instances drawn from \JWTotalImages{} unique images, with \JWTotalRegions{} annotated text regions totaling \JWTotalChars{} characters across \JWUniqueChars{} unique characters.
By design, each task comprises approximately 1,000 instances, allowing for broadly comparable score precision across tasks.
Figure~\ref{fig:sample_images} shows representative images from each task, and Figure~\ref{fig:image_text_properties} visualizes text density and layout properties discussed below.

\paragraph{Text Density and Spatial Layout.} 
Dense STVQA and Receipt KIE are text-dense, often containing several hundred characters per image, whereas Handwriting OCR is intentionally controlled to approximately 100 characters per image (Figure~\ref{fig:char_density}).
Dense STVQA exhibits a long tail beyond 2,000 characters per image, reflecting the high information density of signboards and bulletin boards; this density forces models to locate and integrate information across many text regions, making the task sensitive to both recognition errors and cross-region reasoning failures.
In contrast, the controlled length of Handwriting OCR isolates recognition ability from reasoning, providing a setting in which errors can be attributed almost entirely to character-level reading. 
The spatial distribution of text-region centers (Figure~\ref{fig:text_heatmap}) mirrors these design choices: Dense STVQA regions spread broadly across the frame, Receipt KIE regions form a narrow vertical band consistent with elongated receipt layouts, and Handwriting OCR clusters near the page center along both axes.

\begin{figure}[t]
    \centering
    \includegraphics[width=\linewidth]{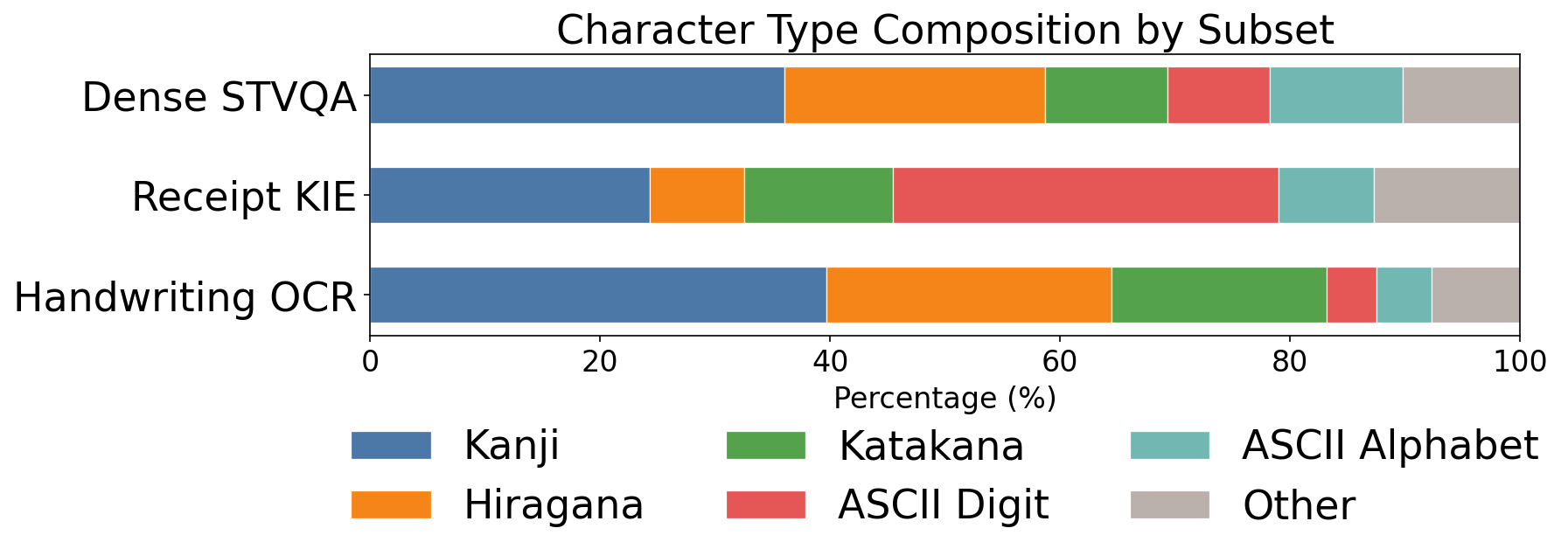}
    \caption{Stacked character-type composition by task. Japanese scripts dominate Dense STVQA and Handwriting OCR, while Receipt KIE allocates a large fraction to ASCII digits.}
    \label{fig:char_distribution}
\end{figure}

\begin{table}[t]
    \centering
    \input{tables/stats_question_types.tex}
\hfill
    \input{tables/stats_receipt_field_fill_rates.tex}
\hfill
    \input{tables/stats_hw_medium_direction.tex}
\end{table}

\paragraph{Character-type Composition.} 
Character-type distributions differ markedly across tasks (Figure~\ref{fig:char_distribution}), which foreshadow the character-type analysis in Section~\ref{sec:analysis}.
Dense STVQA and Handwriting OCR are dominated by kanji (\JWBvqaKanjiPct\% and \JWHwKanjiPct\%) and hiragana (\JWBvqaHiraganaPct\% and \JWHwHiraganaPct\%), consistent with natural Japanese prose and placing heavy demands on mixed-script recognition.
Receipt KIE contains a much larger share of ASCII digits (\JWRkAsciiDigitPct\%), reflecting prices, quantities, and dates; accurate digit reading is therefore a decisive factor for extraction performance in this task.
Overall, \dataset{} contains \JWUniqueKanji{} unique kanji characters.
Of the \JWJoyoKanji{} J\={o}y\={o} kanji (the Japanese government's daily-use list), the dataset covers \JWJoyoCoverage{} (\JWJoyoKanjiCoveragePct\%), and it additionally includes
\JWKanjiBeyondJoyo{} kanji beyond the J\={o}y\={o} set, meaning that models must generalize beyond standard literacy inventories to handle real-world text.

\paragraph{Task-specific properties.}
Each task contributes a distinctive evaluation signal.
In Dense STVQA, questions cover a balanced mix of reasoning types: compositional retrieval, counting, calculation, and spatial reasoning (Table~\ref{tab:question_types}).
Questions are linked to minimal evidence regions, making it possible to separate recognition errors (failing to read individual regions) from reasoning errors (failing to combine correctly read evidence).
In Receipt KIE, field fill rates vary substantially (Table~\ref{tab:receipt_fill_rate}): \texttt{store\_name} and \texttt{date} are present in all receipts, whereas \texttt{store\_address} has the lowest fill rate (\JWRkStoreAddressFillRate\%), reflecting real-world omission patterns and testing whether models can handle missing fields without hallucinating content.
In Handwriting OCR, \JWHwWriters{} writers contributed data with controlled numbers of instances per writer, spanning multiple writing media and directions (Table~\ref{tab:hw_medium}), ensuring that recognition performance is evaluated across a range of handwriting variability rather than being biased toward a single style.

\section{Experiments}
\label{sec:experiments}

\subsection{Experimental Setup}
\label{sec:exp_setup}

\paragraph{Inference.}
We evaluate 14 open-weight VLMs from five model families.
We include four recent high-performing families: Qwen3-VL~\cite{bai-etal-2025-qwen3-vl}, InternVL3.5~\cite{wang-etal-2025-internvl3-5}, Gemma3~\cite{gemmateam-2025-gemma3}, and Phi-4-Multimodal~\cite{abouelenin-etal-2025-phi4mm}.
For families offering multiple sizes, we select variants with 1B to 38B parameters (see Table~\ref{tab:jawildtext_results} for the complete list).
We additionally include Sarashina2.2-Vision~\cite{sbint-2025-sarashina2-2v}, a model built on a Japanese-centric LLM backbone and trained with Japanese document and OCR data.
This selection lets us examine how model scale and language-specific training influence performance on \dataset{}.
We set the temperature to 0 and the maximum token length to 2,048.
Each instance uses a single image at its original resolution; resizing or tiling is applied according to the default preprocessing.

\paragraph{Evaluation.}
To reliably extract the final answer from model output, we enforce machine-parseable output formats using a fixed prompt template for each task.
Models must enclose the final answer in \texttt{\textbackslash boxed\{...\}} for Dense STVQA, return a single JSON object following the predefined schema for Receipt KIE, and output plain text transcriptions for Handwriting OCR.
Any output that cannot be parsed receives a score of~0.
Before scoring, we apply Unicode NFKC normalization to both predictions and references to absorb superficial character-form differences.
In the Dense STVQA task, answers are open-ended and may vary due to differences in units or paraphrasing.
Thus, we adopt judge-based accuracy: an LLM verifier compares each prediction against the reference and returns a binary correctness label.\footnote{We employ \texttt{openai/gpt-5.1-2025-11-13} via the Azure OpenAI API as the judge model. We will release the verifier prompt to reproduce scoring.}
Combined with the evidence region annotations, this binary signal enables error analysis to attribute each failure to recognition or reasoning.
In the Receipt KIE task, outputs are structured, and field boundaries are well-defined. We report overall F1 and field-level accuracy for major header fields. This indicates which field types are most challenging to extract.
For Handwriting OCR, we compute character-level similarity as $\max(0,\; 1 - \mathrm{CER})$, where CER is defined as the Levenshtein distance between prediction and reference, divided by reference length.
Character-level scoring is suitable for Japanese because it lacks explicit word boundaries.
We further report script-type breakdown (e.g., kanji vs.\ hiragana) in CER.
We compute the overall score as the unweighted average of the three task scores.

\subsection{Main Results}
\label{sec:main_results}

\input{tables/main_results.tex}

\begin{figure}[t]
\centering
\includegraphics[width=\linewidth]{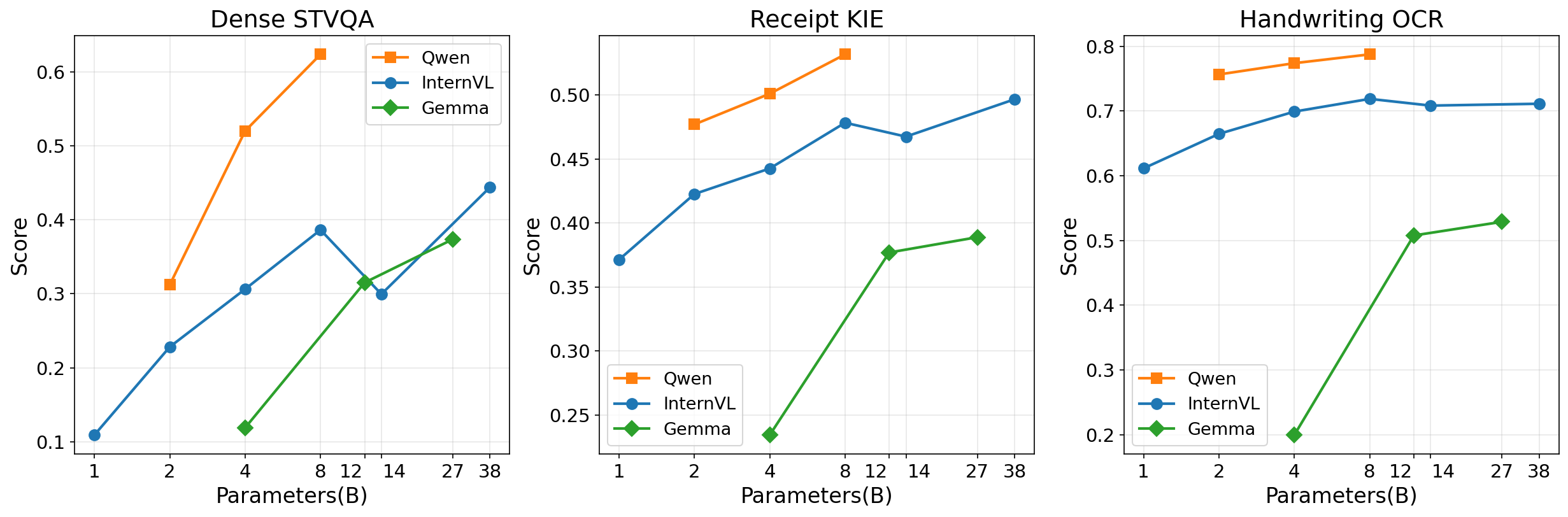}
\caption{Performance scaling trends with model size across benchmark tasks. Larger models improve overall performance, but gains differ by task family.}
\label{fig:scaling_curves}
\end{figure}

Table~\ref{tab:jawildtext_results} summarizes the performance of all evaluated models on \dataset{}.
The best model, Qwen3-VL-8B, achieves an overall score of only 0.64, indicating that Japanese scene text understanding remains a substantial challenge for current open-weight VLMs.
Dense STVQA exhibits the widest performance spread across models (0.008--0.62), suggesting that this task effectively differentiates models with varying levels of Japanese scene text capability.
In Handwriting OCR, 10 of 14 models score at least 0.60, indicating that most models already possess a baseline handwriting recognition ability, though a ceiling around 0.80 persists even for the strongest models.
Performance differences across model families are significant: Qwen3-VL consistently outperforms InternVL3.5 at similar parameter scales, surpassing InternVL3.5 by 0.11 Overall (0.64 vs.\ 0.53) at 8B parameters.
Gemma3 trails both families despite having up to $\sim3.4\times$ more parameters than the best-performing model. At the lower end, Phi-4-Multimodal attains near-zero accuracy on Dense STVQA (0.008), struggling to follow the required output format.
Notably, raw parameter count does not fully explain these gaps.
Sarashina2.2-Vision-3B achieves 0.44 accuracy on Dense STVQA, matching InternVL3.5-38B despite fewer parameters. However, this advantage does not generalize to Receipt KIE or Handwriting OCR, where Sarashina2.2-Vision-3B is comparable to InternVL3.5-2B.
This contrast suggests that the benefit of Japanese-centric training data may be task-dependent.

To examine whether each task captures scaling behavior effectively, we plot performance trends within model families as parameter count (Figure~\ref{fig:scaling_curves}).
Within each family, all three tasks show improvement with scale, but their trajectories differ.
Dense STVQA and Receipt KIE continue to improve across the parameter range we evaluate, with no clear saturation point.
Handwriting OCR, by contrast, plateaus beyond a certain scale within each model family: InternVL3.5 saturates around 0.70 from 4B onward, and Qwen3-VL shows a marginal gain from 2B to 8B (0.76$\rightarrow$0.79).
Across families, however, scale is not decisive: Qwen3-VL-8B surpasses InternVL3.5-38B by 0.09 Overall, indicating that architecture and training data composition can matter more than parameter count alone.

\input{tables/receipt_kie_fields}

The header-field accuracies in Table~\ref{tab:jawildtext_receipt_kie} exhibit sharp variation across field types in Receipt KIE.
Format-constrained fields such as \texttt{time} are relatively well handled, with several top models achieving accuracy above 0.90.
In contrast, \texttt{store\_name} and \texttt{store\_address} remain difficult, with best accuracies of only 0.16 and 0.55, respectively; these fields often require aggregating non-contiguous text spans across the receipt layout rather than copying a single contiguous line.
This gap indicates that the primary bottleneck in Receipt KIE is not character recognition alone but spatial reasoning over document layout.

\begin{figure}[t]
\centering
\begin{minipage}[t]{0.55\linewidth}
\centering
\includegraphics[width=\linewidth]{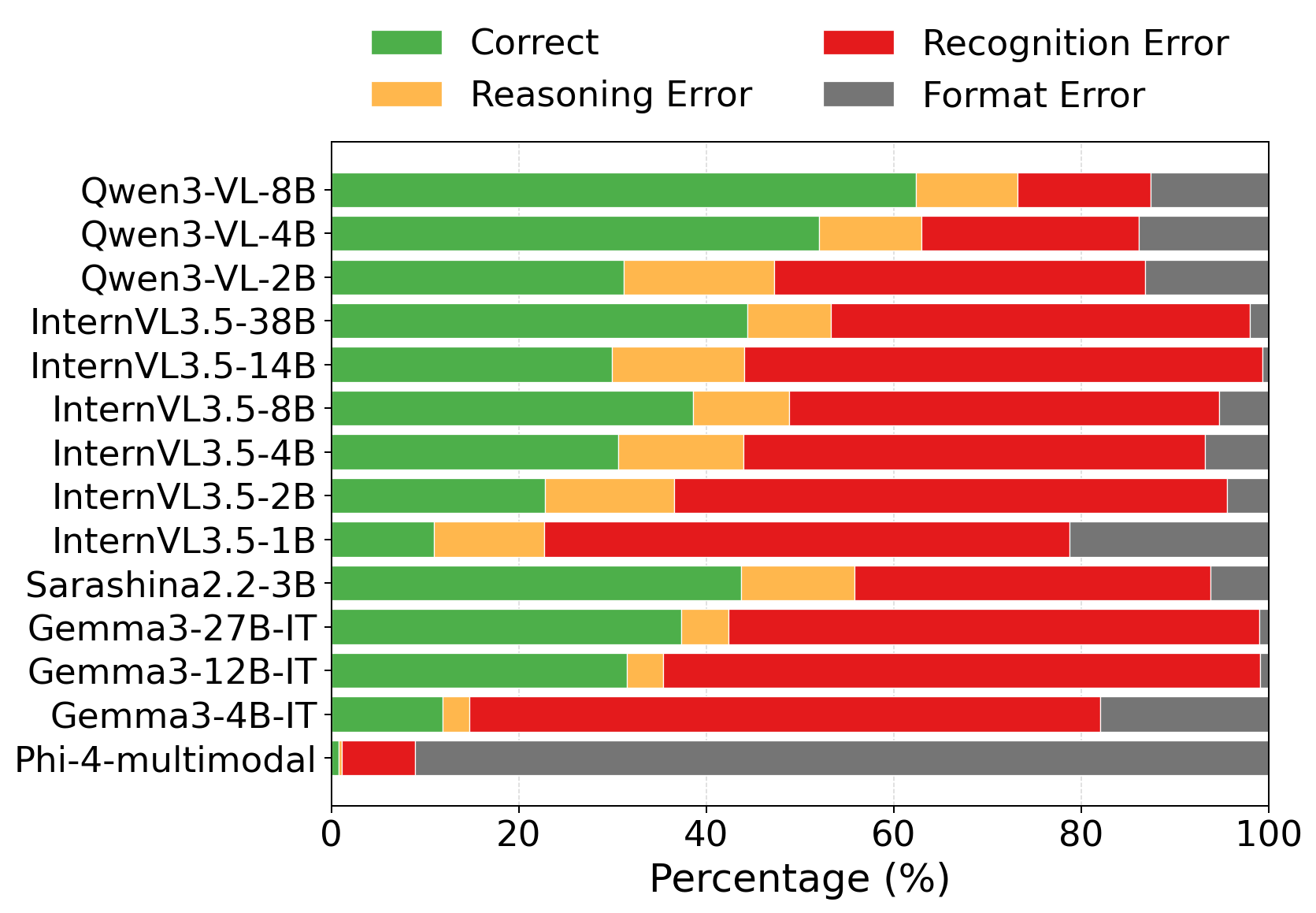}
\captionof{figure}{Error taxonomy on Dense STVQA. Each bar decomposes instances into Correct, Reasoning Error, Recognition Error, and Format Error.}
\label{fig:error_taxonomy}
\end{minipage}
\hfill
\begin{minipage}[t]{0.43\linewidth}
\centering
\includegraphics[width=\linewidth]{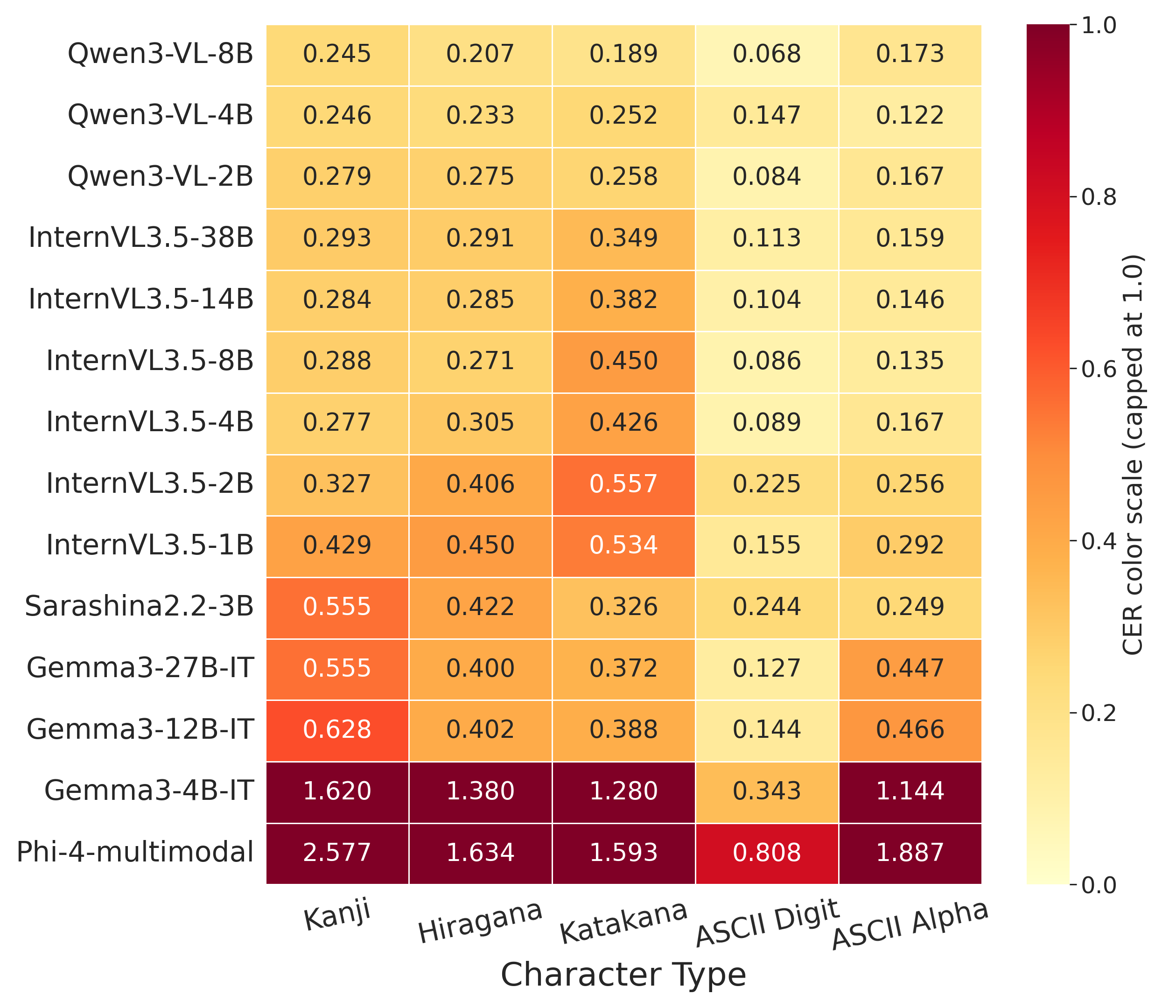}
\captionof{figure}{Character-type CER on Handwriting OCR. Color scale is capped at 1.0; values exceeding 1.0 indicate hallucination-dominant outputs.}
\label{fig:cer-heatmap}
\end{minipage}
\end{figure}

\subsection{Analysis}
\label{sec:analysis}

\paragraph{Error taxonomy for Dense STVQA.}
To disentangle recognition failures from reasoning failures on Dense STVQA, we define an error taxonomy with four categories.
For each Dense STVQA image, we separately prompt the model to transcribe all visible text, independently of the QA task. 
We then compare the resulting transcript against the ground-truth transcriptions of each question's annotated evidence regions: an evidence region is considered ``read'' if its whole ground-truth string appears as an exact substring in the transcript.
Based on this comparison, each instance is assigned one of four outcomes. 
\textbf{Recognition Error}: the answer is not recoverable from the recognized text alone because at least one required evidence region is missing from the transcript. 
\textbf{Reasoning Error}: all evidence regions are present, but the final answer is incorrect. 
\textbf{Format Error}: the output cannot be parsed under the prescribed \verb|\boxed{}| format. 
\textbf{Correct}: the parsed answer matches the reference.

Figure~\ref{fig:error_taxonomy} illustrates that Recognition Error is the largest category for most models, indicating that Japanese scene text recognition remains the primary bottleneck.
Qwen3-VL-8B, which achieves the highest Correct rate (62.3\%), still exhibits a 14.2\% Recognition Error rate, showing that even the best-performing model has not fully overcome this bottleneck.
Sarashina2.2-Vision-3B shows a relatively low Recognition Error rate (38.0\%) compared to other models, such as InternVL3.5-8B (45.9\%) and Gemma3-27B-IT (56.7\%), suggesting that Japanese-centric training may partially improve scene text recognition capability.
InternVL3.5 and Gemma3 remain heavily dominated by Recognition Errors (44.7--58.9\% and 56.7--67.3\%, respectively). 
Phi-4-Multimodal is instead dominated by Format Errors, failing to follow the prescribed output format.
This taxonomy makes visible the failure stage that aggregate accuracy alone cannot reveal, enabling targeted diagnosis of each model's bottleneck.

\begin{figure}[t]
\centering
\includegraphics[width=\linewidth]{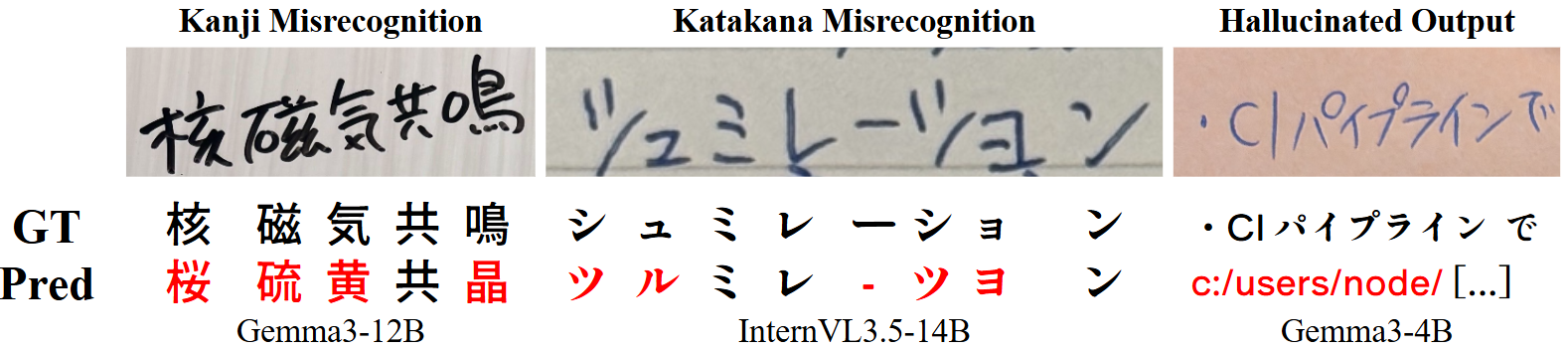}
\caption{Representative failure cases on Handwriting OCR. (Left) Gemma3-12B misrecognizes kanji characters, substituting visually dissimilar characters. (Center) InternVL3.5-14B confuses visually similar katakana characters. (Right) Gemma3-4B produces hallucinated output entirely unrelated to the input image. Red bold text indicates erroneous characters in the model predictions.}
\label{fig:handwriting-failure-examples}
\end{figure}

\paragraph{Script-category analysis on Handwriting OCR.}
To examine how error rates differ across script categories, we decompose CER by script category (kanji, hiragana, katakana, ASCII digits, and ASCII letters).
For each instance, we obtain a character-level alignment
between prediction and reference via minimum edit distance backtracing, then compute CER separately for each category.

Figure~\ref{fig:cer-heatmap} presents per-category CER for each model, showing that CER varies substantially across script categories.
ASCII digits achieve the lowest CER across models, consistent with their small and visually distinct character set. 
Kanji exhibits the highest error rate, which we attribute primarily to the large character inventory: models must disambiguate among thousands of classes, many with limited per-class training exposure.
Since kanji accounts for 39.7\% of all reference characters (Figure~\ref{fig:char_distribution}), its high CER is the dominant contributor to overall scores.

On the other hand, InternVL3.5 models exhibit elevated katakana CER, whereas Gemma3 shows high CER on both kanji and ASCII letters. 
Figure~\ref{fig:handwriting-failure-examples} illustrates these contrasts: InternVL3.5-14B confuses visually similar katakana pairs, while Gemma3-12B misrecognizes kanji with other unrelated kanji characters. 
These differences likely reflect variation in Japanese script coverage during pretraining. Among the weakest models, Gemma3-4B-IT and Phi-4-Multimodal produce CER values exceeding 1.0, indicating that their edit distances exceed the reference length. 
As with the Dense STVQA error taxonomy, stratifying evaluation by linguistically meaningful categories reveals distinct failure profiles that aggregate scoring would obscure.

\input{tables/ocr_specialized_handwriting_results}

\paragraph{Comparison with OCR-specialized models.}
To situate VLM performance on the recognition-dominant Handwriting OCR task, we compare against three OCR-specialized models: DeepSeek-OCR~\cite{deepseek-ocr-2025}, PaddleOCR-VL~\cite{paddleocr-vl-2025}, and olmOCR-2-7B~\cite{olmocr2-2025}, evaluated under the same conditions (Section~\ref{sec:exp_setup}). 
As Table~\ref{tab:ocr_specialized_handwriting_results} shows, the best OCR-specialized model (olmOCR-2-7B, 0.74) falls below the best general-purpose VLM (Qwen3-VL-8B, 0.79) at a comparable parameter scale. 
OCR-specialized models are often positioned as strong baselines for document OCR and document parsing. Still, they perform similarly or worse than general-purpose VLMs when recognizing handwritten text in real-world environments, where diverse writing media, writing instruments, and imaging conditions differ substantially from scanned documents.
This result underscores that robust recognition of handwritten scene text remains an open challenge that cannot be addressed by OCR-specific training alone.

\section{Conclusion}

We introduced \dataset{}, a diagnostic benchmark for evaluating VLMs on Japanese scene text understanding across three complementary tasks: Dense STVQA, Receipt KIE, and Handwriting OCR. 
Benchmarking 14 open-weight VLMs shows that the best model achieves only 0.64 on our unified score, confirming that robust Japanese text understanding in the wild remains far from solved.
Stratifying errors by type and script category reveals that recognition remains the dominant bottleneck, with kanji posing a particular challenge.
Closing the remaining gap will require targeted interventions at each stage, informed by the kind of fine-grained diagnosis that \dataset{} provides.
We hope that \dataset{} will encourage diagnostic scene text evaluation in other typologically diverse languages.

\bibliographystyle{splncs04}
\bibliography{custom}
\end{document}

%% file: tables/dataset_vars.tex
\newcommand{\JWTotalSamples}{3,241}
\newcommand{\JWTotalImages}{2,961}
\newcommand{\JWTotalRegions}{95,705}
\newcommand{\JWTotalChars}{1,117,514}

\newcommand{\JWUniqueChars}{3,643}
\newcommand{\JWBvqaSamples}{1,025}
\newcommand{\JWBvqaImages}{745}
\newcommand{\JWBvqaRegions}{33,608}
\newcommand{\JWBvqaChars}{571,979}
\newcommand{\JWBvqaUniqueChars}{3,313}
\newcommand{\JWBvqaRegionsPerImage}{45.1}

\newcommand{\JWRkSamples}{1,151}
\newcommand{\JWRkImages}{1,151}
\newcommand{\JWRkRegions}{56,095}
\newcommand{\JWRkChars}{433,558}
\newcommand{\JWRkUniqueChars}{2,001}

\newcommand{\JWRkStoreNameFillRate}{100.0}
\newcommand{\JWRkDateFillRate}{100.0}
\newcommand{\JWRkTotalAmountFillRate}{99.8}
\newcommand{\JWRkTimeFillRate}{98.3}
\newcommand{\JWRkTaxAmountFillRate}{95.2}
\newcommand{\JWRkReceiptIdFillRate}{92.0}
\newcommand{\JWRkStoreAddressFillRate}{48.9}
\newcommand{\JWHwSamples}{1,065}
\newcommand{\JWHwImages}{1,065}
\newcommand{\JWHwRegions}{6,002}
\newcommand{\JWHwChars}{111,977}

\newcommand{\JWHwUniqueChars}{1,830}
\newcommand{\JWHwWriters}{51}

\newcommand{\JWHwLined}{470 (43.9\%)}
\newcommand{\JWHwUnlined}{344 (32.3\%)}
\newcommand{\JWHwTablet}{130 (12.2\%)}
\newcommand{\JWHwWhiteboard}{121 (11.4\%)}
\newcommand{\JWHwHorizontal}{4,222 (70.3\%)}
\newcommand{\JWHwVertical}{1,780 (29.7\%)}

\newcommand{\JWBvqaKanjiPct}{36.0}
\newcommand{\JWBvqaHiraganaPct}{22.6}
\newcommand{\JWHwKanjiPct}{39.7}
\newcommand{\JWHwHiraganaPct}{24.8}
\newcommand{\JWRkAsciiDigitPct}{33.5}
\newcommand{\JWUniqueKanji}{2,866}
\newcommand{\JWJoyoKanji}{2,136}

\newcommand{\JWJoyoKanjiCoveragePct}{92.9}
\newcommand{\JWJoyoCoverage}{1,985}
\newcommand{\JWKanjiBeyondJoyo}{881}
\newcommand{\JWBvqaQTypeCompRetr}{406 (39.6\%)}
\newcommand{\JWBvqaQTypeCount}{374 (36.5\%)}
\newcommand{\JWBvqaQTypeCalc}{125 (12.2\%)}
\newcommand{\JWBvqaQTypeSpatial}{84 (8.2\%)}
\newcommand{\JWBvqaQTypeOther}{36 (3.5\%)}

%% file: tables/stats_overall.tex
\begin{table}[t]
\centering
\footnotesize
\caption{Summary statistics of \dataset. 
\textbf{\#Instances} denotes the evaluation unit: an instance is a question–answer pair in Dense STVQA, and a single image in Receipt KIE and Handwriting OCR. 
\textbf{\#Regions} denotes the number of annotated quadrilateral text regions.
}
\label{tab:dataset_stats}
\setlength{\tabcolsep}{2pt}
\begin{tabular}{lrrrrr}
\toprule
\textbf{Subset} & \textbf{\#Instances} & \textbf{\#Images} & \textbf{\#Regions} & \textbf{\#Chars} & \textbf{\#Unique Chars} \\
\midrule
Dense STVQA     & \JWBvqaSamples & \JWBvqaImages & \JWBvqaRegions & \JWBvqaChars & \JWBvqaUniqueChars \\
Receipt KIE     & \JWRkSamples & \JWRkImages & \JWRkRegions & \JWRkChars & \JWRkUniqueChars \\
Handwriting OCR & \JWHwSamples & \JWHwImages & \JWHwRegions & \JWHwChars & \JWHwUniqueChars \\
\midrule
\textbf{Total}  & \textbf{\JWTotalSamples} & \textbf{\JWTotalImages} & \textbf{\JWTotalRegions} & \textbf{\JWTotalChars} & \textbf{\JWUniqueChars} \\
\bottomrule
\end{tabular}
\end{table}

%% file: tables/stats_question_types.tex
\begin{minipage}[t]{0.35\linewidth}
\centering
\scriptsize
\captionof{table}{Question-type distribution in Dense STVQA.}
\label{tab:question_types}
\setlength{\tabcolsep}{1.5pt}
\begin{tabular}{@{}p{0.56\linewidth}r@{}}
\toprule
\textbf{Question type} & \textbf{\# (\%)} \\
\midrule
Compositional retrieval & \JWBvqaQTypeCompRetr \\
Counting           & \JWBvqaQTypeCount \\
Calculation        & \JWBvqaQTypeCalc \\
Spatial            & \JWBvqaQTypeSpatial \\
Other              & \JWBvqaQTypeOther \\
\bottomrule
\end{tabular}
\end{minipage}

%% file: tables/stats_receipt_field_fill_rates.tex
\begin{minipage}[t]{0.25\linewidth}
\centering
\scriptsize
\captionof{table}{Receipt KIE field fill rates.}
\label{tab:receipt_fill_rate}
\setlength{\tabcolsep}{1.5pt}
\begin{tabular}{@{}p{0.45\linewidth}r@{}}
\toprule
\textbf{Field} & \textbf{Fill Rate} \\
\midrule
\texttt{store\_name} & \JWRkStoreNameFillRate\% \\
\texttt{date} & \JWRkDateFillRate\% \\
\texttt{total\_amount} & \JWRkTotalAmountFillRate\% \\
\texttt{time} & \JWRkTimeFillRate\% \\
\texttt{tax\_amount} & \JWRkTaxAmountFillRate\% \\
\texttt{receipt\_id} & \JWRkReceiptIdFillRate\% \\
\texttt{store\_address} & \JWRkStoreAddressFillRate\% \\
\bottomrule
\end{tabular}
\end{minipage}

%% file: tables/stats_hw_medium_direction.tex
\begin{minipage}[t]{0.34\linewidth}
\centering
\scriptsize
\captionof{table}{Writing surface and direction in Handwriting OCR.}
\label{tab:hw_medium}
\setlength{\tabcolsep}{1.5pt}
\begin{tabular}{@{}p{0.45\linewidth}r@{}}
\toprule
\textit{Medium} & \textit{\#Instances (\%)} \\
\midrule
Lined paper   & \JWHwLined \\
Unlined paper & \JWHwUnlined \\
Tablet        & \JWHwTablet \\
Whiteboard    & \JWHwWhiteboard \\
\midrule
\textit{Direction}  & \textit{\#Regions (\%)} \\
\midrule
Horizontal & \JWHwHorizontal \\
Vertical   & \JWHwVertical \\
\bottomrule
\end{tabular}
\end{minipage}

%% file: tables/main_results.tex
\begin{table}[t]
\centering
\caption{Results on the JaWildText benchmark.}
\label{tab:jawildtext_results}
\resizebox{\textwidth}{!}{%
    \begin{tabular}{lr|cccc}
    \toprule
    Model & Params & Overall & Dense STVQA & Receipt KIE & Handwriting OCR \\
     &  &  & Accuracy & F1 & 1-CER \\
    \midrule
    Qwen3-VL-8B & 8B & \textbf{0.64} & \textbf{0.62} & \textbf{0.53} & \textbf{0.79} \\
    Qwen3-VL-4B & 4B & 0.60 & 0.52 & 0.50 & 0.77 \\
    Qwen3-VL-2B & 2B & 0.52 & 0.31 & 0.48 & 0.76 \\
    InternVL3.5-38B & 38B & 0.55 & 0.44 & 0.50 & 0.71 \\
    InternVL3.5-14B & 14B & 0.49 & 0.30 & 0.47 & 0.71 \\
    InternVL3.5-8B & 8B & 0.53 & 0.39 & 0.48 & 0.72 \\
    InternVL3.5-4B & 4B & 0.48 & 0.31 & 0.44 & 0.70 \\
    InternVL3.5-2B & 2B & 0.44 & 0.23 & 0.42 & 0.66 \\
    InternVL3.5-1B & 1B & 0.37 & 0.11 & 0.37 & 0.61 \\
    Sarashina2.2-Vision-3B & 3B & 0.50 & 0.44 & 0.40 & 0.68 \\
    Gemma3-27B-IT & 27B & 0.43 & 0.37 & 0.39 & 0.53 \\
    Gemma3-12B-IT & 12B & 0.40 & 0.32 & 0.38 & 0.51 \\
    Gemma3-4B-IT & 4B & 0.19 & 0.12 & 0.23 & 0.20 \\
    Phi-4-Multimodal & 14B & 0.18 & 0.008 & 0.23 & 0.29 \\
    \bottomrule
    \end{tabular}
}
\end{table}

%% file: tables/receipt_kie_fields.tex
\begin{table}[t]
\centering
\caption{Receipt KIE field-level accuracy on JaWildText.}
\label{tab:jawildtext_receipt_kie}
\resizebox{\textwidth}{!}{%
    \begin{tabular}{lr|ccccccc}
    \toprule
    Model & Params & Store Name & Address & Receipt ID & Date & Time & Total & Tax \\
    \midrule
    Qwen3-VL-8B & 8B & 0.14 & 0.52 & 0.13 & \textbf{0.81} & 0.91 & 0.81 & 0.77 \\
    Qwen3-VL-4B & 4B & \textbf{0.16} & \textbf{0.55} & 0.11 & 0.28 & 0.92 & \textbf{0.84} & 0.80 \\
    Qwen3-VL-2B & 2B & 0.11 & 0.43 & 0.08 & 0.52 & 0.88 & 0.82 & \textbf{0.82} \\
    InternVL3.5-38B & 38B & 0.15 & 0.34 & 0.18 & 0.52 & \textbf{0.92} & 0.83 & 0.82 \\
    InternVL3.5-14B & 14B & 0.15 & 0.35 & 0.13 & 0.26 & 0.89 & 0.83 & 0.79 \\
    InternVL3.5-8B & 8B & 0.11 & 0.31 & 0.13 & 0.50 & 0.89 & 0.79 & 0.74 \\
    InternVL3.5-4B & 4B & 0.14 & 0.29 & \textbf{0.21} & 0.26 & 0.90 & 0.77 & 0.73 \\
    InternVL3.5-2B & 2B & 0.13 & 0.30 & 0.11 & 0.24 & 0.89 & 0.77 & 0.75 \\
    InternVL3.5-1B & 1B & 0.09 & 0.23 & 0.10 & 0.33 & 0.80 & 0.69 & 0.70 \\
    Sarashina2.2-Vision-3B & 3B & 0.11 & 0.28 & 0.18 & 0.28 & 0.75 & 0.72 & 0.67 \\
    Gemma3-27B-IT & 27B & 0.11 & 0.20 & 0.08 & 0.52 & 0.71 & 0.72 & 0.64 \\
    Gemma3-12B-IT & 12B & 0.06 & 0.21 & 0.11 & 0.39 & 0.67 & 0.70 & 0.60 \\
    Gemma3-4B-IT & 4B & 0.03 & 0.10 & 0.05 & 0.20 & 0.59 & 0.59 & 0.35 \\
    Phi-4-Multimodal & 14B & 0.01 & 0.004 & 0.07 & 0.13 & 0.51 & 0.51 & 0.48 \\
    \bottomrule
    \end{tabular}
}
\end{table}

%% file: tables/ocr_specialized_handwriting_results.tex
\begin{table}[t]
\centering
\caption{Handwriting OCR results for OCR-specialized models.}
\setlength{\tabcolsep}{4pt}
\label{tab:ocr_specialized_handwriting_results}
\begin{tabular}{lr|c}
\toprule
Model & Params & Handwriting OCR (1-CER)\\
\midrule
olmOCR-2-7B & 7B & \textbf{0.74}\\
PaddleOCR-VL & 0.9B & 0.61\\
DeepSeek-OCR & 3B & 0.53\\
\bottomrule
\end{tabular}
\end{table}